\documentclass[10pt,twocolumn,letterpaper]{article}

\usepackage{wacv}
\usepackage{times}
\usepackage{epsfig}
\usepackage{graphicx}
\usepackage{amsmath}
\usepackage{amssymb}
\usepackage{booktabs}
\usepackage{tikz}
\usepackage{comment}
\usepackage{amssymb}
\usepackage{color}
\usepackage{caption}
\usepackage{mathtools}
\usepackage{subcaption}
\usepackage[accsupp]{axessibility}
\usepackage{authblk}

\def\wacvPaperID{795} 

\wacvapplicationstrack 

\wacvfinalcopy 

\ifwacvfinal
\usepackage[breaklinks=true,bookmarks=false]{hyperref}
\else
\usepackage[pagebackref=true,breaklinks=true,colorlinks,bookmarks=false]{hyperref}
\fi

\begin{document}

\title{Cross-View Image Sequence Geo-localization}

\author[1]{Xiaohan Zhang}
\author[2]{Waqas Sultani}
\author[1]{Safwan Wshah}
\affil[1]{Department of Computer Science, University of Vermont, USA }
\affil[2]{Intelligent Machine Lab, Information Technology University, Pakistan \authorcr {\tt \{Xiaohan.Zhang, Safwan.Wshah\}@uvm.edu waqas.sultani@itu.edu.pk}}


\maketitle
\thispagestyle{empty}

\begin{abstract}
Cross-view geo-localization aims to estimate the GPS location of a query ground-view image by matching it to images from a reference database of geo-tagged aerial images. To address this challenging problem, recent approaches use panoramic ground-view images to increase the range of visibility. Although appealing, panoramic images are not readily available compared to the videos of limited Field-Of-View (FOV) images. In this paper, we present the first cross-view geo-localization method that works on a sequence of limited FOV images. Our model is trained end-to-end to capture the temporal structure that lies within the frames using the attention-based temporal feature aggregation module. To robustly tackle different sequences length and GPS noises during inference, we propose to use a sequential dropout scheme to simulate variant length sequences. To evaluate the proposed approach in realistic settings, we present a new large-scale dataset containing ground-view sequences along with the corresponding aerial-view images. Extensive experiments and comparisons demonstrate the superiority of the proposed approach compared to several competitive baselines.


\end{abstract}
\section{Introduction}
\label{sec:intro}

Cross-view image geo-localization aims to determine the geospatial location from where an image was taken (also known as the query image) in a database of geo-tagged aerial images (also known as reference images) ~\cite{CVUSA,SAFA,lin2013cross,Vigor}. Estimating geo-spatial locations from images has many important applications such as autonomous driving~\cite{UAV1}, robot navigation\cite{biswas2011depth,kim2017satellite}, augmented reality (AR)~\cite{AR1}, and unmanned aerial vehicle (UAV) navigation~\cite{UAV1}.

Despite the huge research efforts that have been done on this problem, image geo-localization remains far from being solved and is considered one of the most challenging tasks in the computer vision field due to: 1) the drastic appearance differences between the query images and reference images, 2) capturing time gaps between the query image and the reference image results in different illumination conditions, weather, and objects and, 3) differences in resolution at which ground and aerial images are captured.

\begin{figure}[!t]
    \centering
    \includegraphics[width=0.45\textwidth]{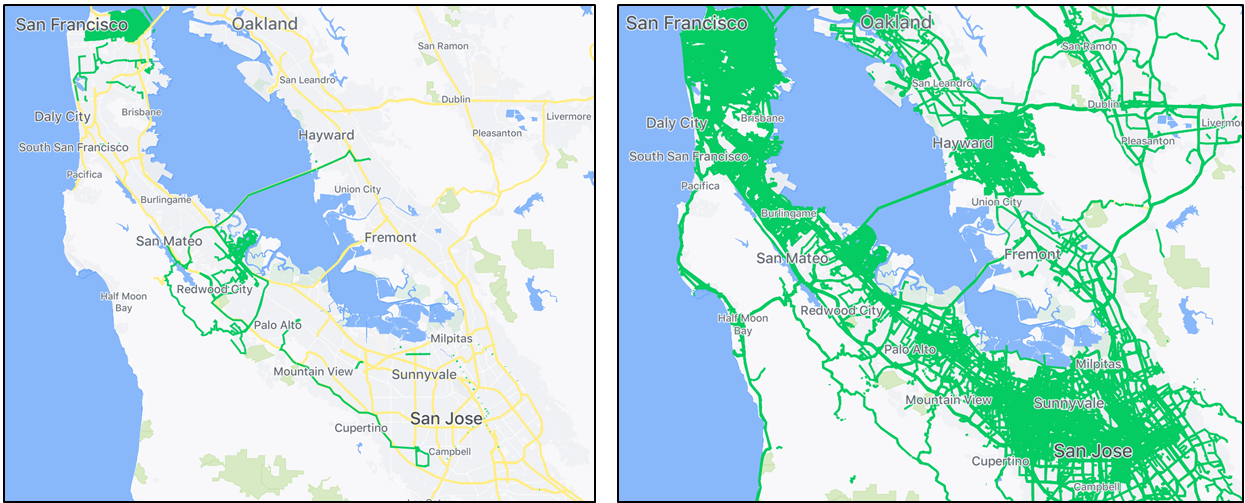}
    \caption{Comparison of the coverage area (green lines) of user uploaded street view images between panoramas (left) and limited FOV images (right) in San Francisco, USA from Mapillary~\cite{mapillary}.}
    \label{fig:pano_cover}
\end{figure}

Recent research in cross-view image geo-localization has shown tremendous progress on large-scale datasets~\cite{CVUSA,liu2019lending,Vigor}, but they heavily rely on panoramic query images~\cite{CVUSA,cvmnet,SAFA,bridging,featureTransport,hardTriplet,liu2019lending,revisiting,Vigor,survey}. Even though panoramic images provide richer contextual information than normal limited Field-Of-View (FOV) images, in practice, limited FOV images are more common and easier to capture from smartphones, dash cams, and digital single-lens reflex (DSLR) cameras.
Fig.~\ref{fig:pano_cover} shows the comparison of coverage area of users uploaded street view images between panoramas and limited FOV images in San Francisco, USA on Mapillary~\cite{mapillary}. Moreover, even map platforms such as Google Street View (GSV) provides panoramas only for a few historic or tourist attraction places for several countries such as China, Qatar, and Pakistan. However, limited FOV street view images are available across 190 countries in the most of the regions as shown on Mapillary~\cite{mapillary}. Clearly, the limited FOV images are much more popular than panoramic images. This applies to all other countries and is more noticeable in developing countries where for the most part, panoramic images are not available at all.

Due to the recent advancement of autonomous vehicles and the Advanced Driving Assistance System (ADAS), frontal street view videos are easily accessible from the dash cams in current vehicles. Instead of using unpopular panoramic images~\cite{DSM,comingDown,SAFA}, expanding cross-view geo-localization algorithms to work on sequences of images is more practical and more acceptable in real-world scenarios. On the other hand, current cross-view geo-localization approaches~\cite{DSM,comingDown,SAFA,cvmnet,featureTransport,Vigor,Vo} deal mainly with a single image for geo-localization and cannot be used directly to capture the temporal structure that lies within a sequence of FOV frames. Thus, it is a natural extension to expand cross-view geo-localization methods on sequences of limited FOV images named cross-view image sequence geo-localization.

In this paper, a new cross-view geo-localization approach is proposed that works on sequences of limited FOV images. Our model is trained end-to-end to capture \textit{temporal feature representations} that lie within the images for better geo-localization. Although our model is trained on fixed-length temporal sequence, it tackles the challenge of variable length sequence during the inference phase through a novel sequential dropout scheme. To the best of our knowledge, we are the first one to propose end-to-end cross-view geo-localization from \textit{sequences} of images. We refer to this task as \textit{cross-view image sequence geo-localization}. Furthermore, to facilitate future research in cross-view geo-localization from sequences, we put forward a new dataset and compare our proposed model with several recent baselines. In summary, our main contributions are as follows: \\
1) We propose a new end-to-end approach, \textit{cross-view image sequence} \textit{geo-localization}, that geo-localizes a query sequence of limited FOV ground images and its corresponding aerial images.\\
2) We introduce the first large-scale cross-view image sequence geo-localization dataset.\\
3) We propose a novel temporal feature aggregation technique that learns an end-to-end feature representation from a sequence of limited FOV images for sequence geo-localization.\\
4) We propose a new sequence dropout method to predict coherent features on sequences of different lengths. The proposed dropout method helps in regularizing our model and achieve more robust results.

\section{Related Work}
\label{sec:related_works}
\noindent\textbf{Cross-view Image Geo-localization: }Before the deep learning era, cross-view image geo-localization methods were based on hand-crafted features~\cite{lin2013cross,castaldo2015semantic} such as HoG~\cite{HoG}, GIST~\cite{gist}, self-similarity~\cite{selfSim}, and color histograms. Conventional methods struggled with matching accuracy because of the quality of the features.
Due to the resurgence of deep learning in numerous computer vision applications, several deep learning based geo-localization methods~\cite{CVUSA,lin2015learning,Vo} have been proposed to extract features from fined-tuned CNN models to improve the cross-view geo-localization accuracy. More recently, Hu \etal~\cite{cvmnet} proposed to aggregate features by NetVLAD~\cite{netvlad} layer which achieved significant performance improvements. Shi \etal~\cite{featureTransport} proposed a feature transport module for aligning features from aerial view and street view images. Liu \etal~\cite{liu2019lending} explored fusing orientation information into the model which boosted performance. With the development of Generative Adversarial Networks (GANs)~\cite{gan}, Regmi \etal~\cite{bridging} proposed a GAN-based cross-view image geo-localization approach using a feature fusion training strategy. Zhu \etal~\cite{Vigor} recently proposed a new approach (VIGOR) that does not require a one-to-one correspondence between ground images and aerial images. It is also worth mentioning that some methods~\cite{SAFA,DSM,comingDown} based on ground-level panorama employ the polar transformation which bridges the domain gap between reference images and query images by prior geometric knowledge. By leveraging this prior geometric property, Shi \etal~\cite{SAFA} proposed Spatial Aware Feature Aggregation (SAFA) which improves the results on CVUSA~\cite{CVUSA} and CVACT~\cite{liu2019lending} by a large margin. Similar to \cite{bridging}, Toker \etal~\cite{comingDown} combined SAFA~\cite{SAFA} with a GAN. Their proposed method achieved state-of-the-art results on CVUSA~\cite{CVUSA} and CVACT~\cite{liu2019lending}. However, to perform the polar transformation, the query image is assumed to be aligned at the center of its reference aerial image which is not always guaranteed in real-world scenarios. The above-mentioned methods rely on panoramic ground-level images. By contrast, our method used more easily available limited FOV images.

We noticed that some previous works~\cite{DSM,Vo,buildingMatch} studied the cross-view image geo-localization problem using a single limited FOV image as a query. Tian \etal~\cite{buildingMatch} proposed a graph-based method that matches the detected buildings in both ground images and aerial images. This method was only applicable in metropolitan areas which contain dense buildings. DBL~\cite{Vo} proposed by Vo \textit{et al} focused on geo-localizing the scene in the image rather than the location of the camera. Dynamic Similarity Matching proposed by Shi \etal~\cite{DSM} 
required polar transformed aerial images as input. Compared to these methods, we assume neither aligned ground-level images nor that our method only works in metropolitan areas. Furthermore, instead of geo-localizing a single limited FOV image, our approach geo-localizes a sequence of limited FOV images. 

Recently, Regmi and Shah~\cite{Regmi_2021_ICCV} proposed to geo-localize video sequences in the same-view setting by using a geo-temporal feature learning network and a trajectory smoothing network. On the other hand, in this paper, we incorporate aerial images and ground video sequences to address the problem of cross-view image sequence geo-localization by proposing a transformer-based model. Current cross-view geo-localization approaches can be used for sequential cross-view geo-localization trivially by applying them frame by frame as proposed in ~\cite{kim2017satellite}. However, we propose an end-to-end approach that automatically processes a whole sequence of images and correlates their features with the corresponding aerial image by building a better feature representation in both temporal and spatial domains. We have compared our results with the best models in the literature that could be applied to our dataset as discussed in the experiments section.


\noindent\textbf{Transformer/multi-head attention: }
Recently, Vaswani \etal~\cite{attention} proposed the transformer module and demonstrated its ability in catching the temporal correlation in time series data. Using the transformer, several works~\cite{roberta,GPT,bert} achieved remarkable results in natural language processing tasks. In computer vision, transformers have been used for image classification~\cite{ViT}, video segmentation~\cite{segmentationTrans}, object detection~\cite{DETR}, and same-view video geo-localization~\cite{Regmi_2021_ICCV}. In this paper, we combined the transformer with the cross-view image sequence geo-localization to effectively utilize the full range of visibility from the sequential data. Our experiments showed that the transformer can learn to fuse and summarize several features from a sequence of images and predict robust results.
\section{Dataset}
\label{sec:dataset}

\begin{table*}[!t]
\footnotesize
\centering
\begin{tabular}{c|ccccccccccccccccc}
\toprule
\textbf{Dataset Comparison}& Vo~\cite{Vo}             & \multicolumn{4}{c}{CVACT~\cite{liu2019lending}}    & \multicolumn{4}{c}{CVUSA~\cite{CVUSA}}   & \multicolumn{4}{c}{VIGOR~\cite{Vigor}}    & \multicolumn{4}{c}{Ours}      \\ \midrule
\# of Aerial Images          & $>$ 1M & \multicolumn{4}{c}{128, 334} & \multicolumn{4}{c}{44,416} & \multicolumn{4}{c}{90,618}  & \multicolumn{4}{c}{38,863}   \\
\# of Ground-level Images       & $>$ 1M & \multicolumn{4}{c}{128, 334} & \multicolumn{4}{c}{44, 416} & \multicolumn{4}{c}{105, 214} & \multicolumn{4}{c}{118, 549}  \\
\# of Ground Images per aerial image           & 1             & \multicolumn{4}{c}{1}       & \multicolumn{4}{c}{1}      & \multicolumn{4}{c}{$\sim$5}      & \multicolumn{4}{c}{$\sim$7}       \\
Coverage            & Urban, Suburb             & \multicolumn{4}{c}{Urban, Suburb}       & \multicolumn{4}{c}{Urban, Suburb}      & \multicolumn{4}{c}{Urban}      & \multicolumn{4}{c}{Urban, Suburb}       \\
Seamless Sampling            & No             & \multicolumn{4}{c}{No}       & \multicolumn{4}{c}{No}      & \multicolumn{4}{c}{Yes}      & \multicolumn{4}{c}{Yes}       \\
Sequential Ground-level Images & No             & \multicolumn{4}{c}{No}       & \multicolumn{4}{c}{No}      & \multicolumn{4}{c}{No}       & \multicolumn{4}{c}{Yes}       \\
Orientation                    & Yes            & \multicolumn{4}{c}{Yes}      & \multicolumn{4}{c}{Yes}     & \multicolumn{4}{c}{Yes}      & \multicolumn{4}{c}{Yes}       \\
Ground-level GPS Location      & No             & \multicolumn{4}{c}{Same}  & \multicolumn{4}{c}{Same} & \multicolumn{4}{c}{Arbitrary} & \multicolumn{4}{c}{Arbitrary} \\ \bottomrule
\end{tabular}%
\caption{Comparison between our proposed dataset and other existing cross-view image geo-localization datasets.}
\label{tab:DatasetComparision}
\end{table*}

\subsection{Previous Datasets}

Many datasets have been proposed for cross-view image geo-localization~\cite{CVUSA,liu2019lending,Vigor,Vo}. Vo \etal~\cite{Vo} proposed a large-scale cross-view geo-localization dataset consisting of more than 1 million pairs of satellite-ground images. The authors collected aerial images from Google Maps and the corresponding ground images from Google Street View from eleven different US cities. Workman \etal~\cite{CVUSA} proposed a Cross-View USA (CVUSA) dataset containing more than 1 million ground-level images across the whole USA. Later, Zhai \etal~\cite{zhai} refined the CVUSA dataset by pairing $44,416$ aerial-ground images and this has become one of the most popular datasets in this field. In this paper, we refer to this refined version as CVUSA.
CVACT~\cite{liu2019lending} followed the same structure as CVUSA and had the same number of training samples as CVUSA but had 10 times more testing pairs. Recently, Zhu \etal~\cite{Vigor} proposed the VIGOR dataset which is the first non one-to-one correspondent cross-view image geo-localization dataset collected randomly from four major US cities. In order to have systems for practical scenarios in which the queries and reference images pairs are not guaranteed to be always perfectly aligned, VIGOR defined `positive' and `semi-positive' ground images in one single aerial image. Note that current cross-view geo-localization datasets cannot be easily converted to sequential dataset. To the best of our knowledge, there is no existing dataset that provides \textit{sequential} ground-level images and their corresponding aerial images for cross-view image geo-localization. 

\subsection{Proposed Dataset}
Since existing cross-view geo-localization datasets~\cite{Vo,CVUSA,liu2019lending,Vigor} contain only discrete ground images, we collected a new cross-view image sequence geo-localization dataset containing limited FOV images which are much more available and applicable for real-world systems. Table \ref{tab:DatasetComparision} demonstrates the comparison of our proposed dataset with the existing cross-view image geo-localization datasets. Below, we first explain the procedures we followed to collect the ground-level images and then describe the process of capturing aerial imagery.

\subsubsection{Ground-Level Imagery}
\label{sec::ground_imagery}
Our data was collected using the Fugro Automatic Road Analyzer (ARAN)\footnote{https://www.fugro.com/our-services/asset-integrity/roadware/equipment-and-software} which is a road data capturing vehicle capable of collecting different data modalities such as image, LiDAR, and pavement laser. ARAN is also equipped with a GPS and an inertial measurement unit (IMU) sensor for providing precise GPS locations and camera poses. The raw dataset contains over $5000$km of urban and suburban roads, and highways in both directions in the state of Vermont, US. In our dataset, we only used the frontal camera images with a resolution of $1920 \times 1080$. The distance between each capture point is approximately $8m$ and the FOV of the camera is around $120^{\circ}$. GPS location and camera heading (compass direction) are also provided for each ground-level image. To represent more real-world scenarios, our dataset contains approximately $70\%$ of images from suburban areas and $30\%$ from urban areas which may be collected from one or two-way driving directions. The ratio of the collected two-way driving direction data is around $30\%$ in which the same street images are captured from both driving directions, for example, north-to-south and south-to-north. The total number of ground images is $118,549$ resulting in $38,863$ aerial pairs as explained in the following sections.
Our dataset covers around $500$ kilometers of roads in Vermont. Please refer to the supplementary material for more information.

\subsubsection{Sequence Formation}

\begin{figure}[!t]
    \centering
    \includegraphics[width=0.45\textwidth]{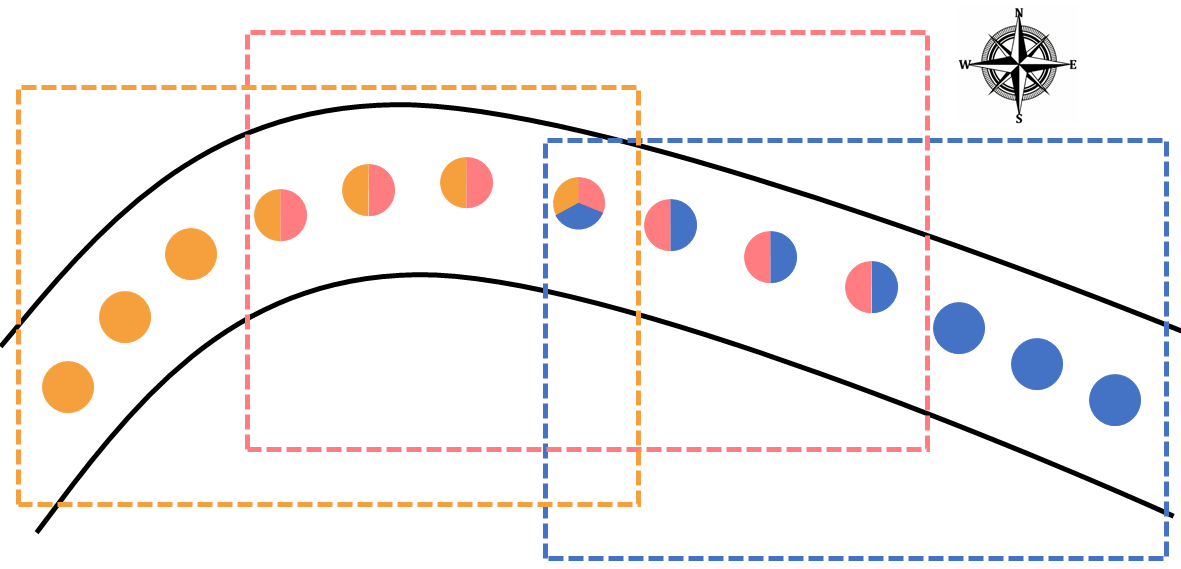}
    \caption{Demonstration of our ground-level images sampling strategy. In this example, three aerial images are captured (yellow, pink , and blue boxes) based on the locations of the ground images (colored circles). Each circle inside these boxes belong to that aerial image. If one circle has multiple colors, it belongs to multiple sequences.}
    \label{fig:sample}
\end{figure}

After obtaining the raw ground-level data as described in section~\ref{sec::ground_imagery}, long sequences of raw data should be segmented into several small sequences to be used for cross-view geo-localization. A simple but effective greedy algorithm is employed. Given the raw ground-level image sequence data as $S={s_0,s_1,...,s_N}$ where $N$ is the number of ground-level images that would be segmented into sequence splits. It is not required to keep the same number of images in each split since in real-world scenarios the number of images in a sequence can be variable due to different hardware or software configurations. However, to perform the retrieval task, it is required that the images of any resulting sequence must \textit{lay within one single aerial image}. Our algorithm iterates through each image in $S$, denoting the first image as $s_0$. After that the distance between $s_0$ and $s_t$ is calculated. If the distance is less than a preset threshold value $\Delta$, we step to the next image $s_{t+1}$. Otherwise, $[s_0,s_t]$ is the segmented sequence. Then the image in the middle of $s_0$ and $s_t$ is set as starting point for the next segment. This process is visualized in Fig.~\ref{fig:sample}. The circles displayed with the same color are segmented in one sequence. If one circle has multiple colors, this circle co-exists in two or more segments. We empirically choose $\Delta = 50\ m$ to ensure that images in any sequence fall within one aerial image with zoom level $20$. To make the training procedure consistent and simple, which will be discussed later, we removed $72$ sequences which contain less than $7$ images. This finally results in $38,863$ sequences with an average of seven images per segment. This sequence formation strategy guarantees that the formed sequence is covered by a single aerial image and does not need to know the length of the raw data. Note that our approach needs seven frames during training. However, we do not have such restrictions while in testing. In real-world scenarios, the distance between each frame may vary and one can simply use techniques such as IMU sensors or visual odometry~\cite{visual_odo} to estimate the distance between frames.

\subsubsection{Aerial Imagery}

\begin{figure}[!t]
    \centering
    \includegraphics[width=0.47\textwidth]{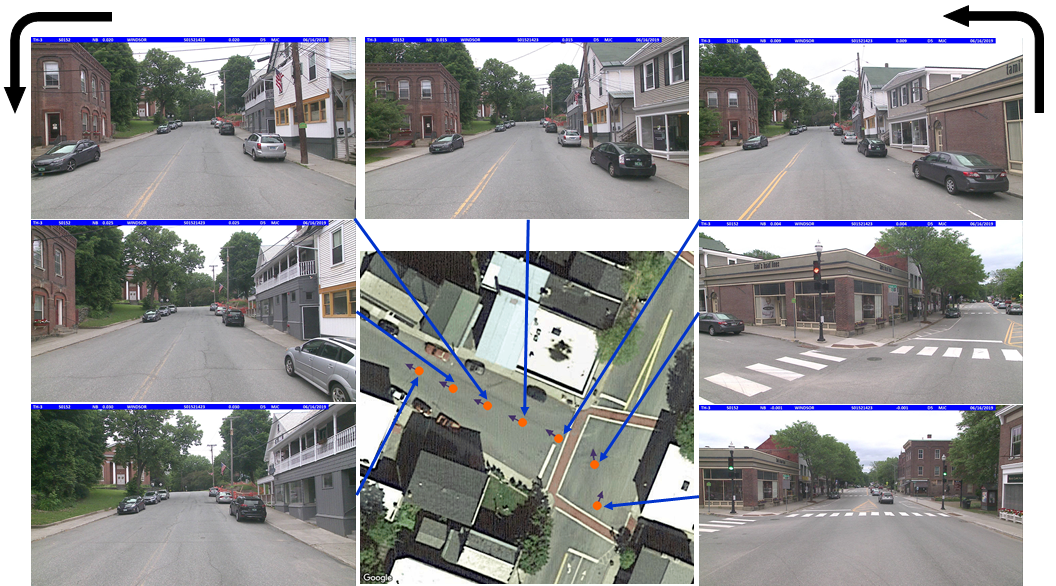}
    \caption{A sequence sample from our dataset. The aerial image is in the center and the ground images are at the edges. Each orange dot represents the location of one ground image indicated by the blue arrow. The grey arrow of each dot represents the heading direction of the camera.}
    \label{fig:dataset}
\end{figure}

Google Maps Static API~\cite{google_static_api} is employed to obtain the aerial images for each sequence. Assuming the ground images in a single sequence are on a planar surface, we can determine the geometric center (the arithmetic mean location) of the aerial image for a given sequence. In this way, the aerial image can cover the whole sequence. A random shift at most $5$ meters is applied to each aerial image to simulate real-world scenarios. This results a one-to-one correspondence between ground sequences and aerial images. The total number of collected aerial images is $38,863$. Following VIGOR~\cite{Vigor}, each aerial image is captured at the zoom level of $20$ with an resolution of $640 \times 640$. The ground resolution is approximately $0.114 \text{m}$ A sample pair of ground-aerial images from our dataset is shown in Fig.~\ref{fig:dataset}.


\section{Proposed Methods}
\label{sec:proposed_method}

\begin{figure*}[!ht]
    \centering
    \includegraphics[width=0.76\textwidth]{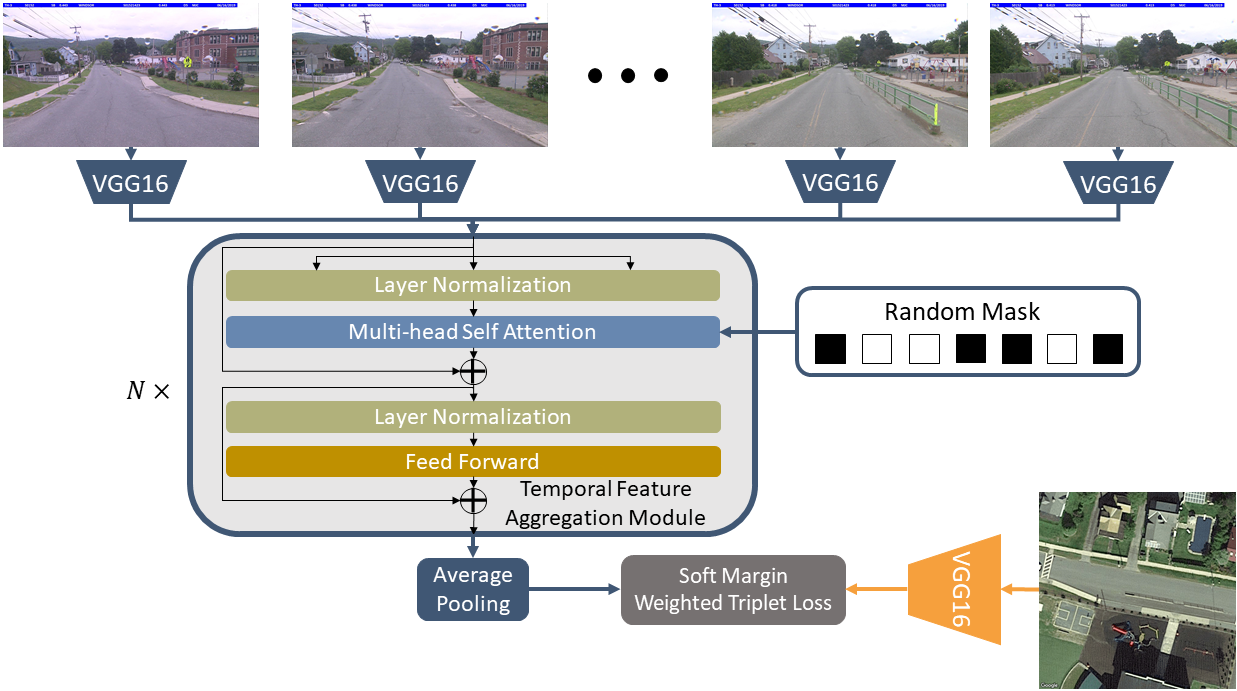}
    \caption{An overview of our proposed method which contains two main parts. The ground features extraction branch (components in dark blue), and the aerial features extraction branch (components in orange). The ground-level features extraction branch takes a sequence of images as input. The aerial features extraction takes the aerial image as input.}
    \label{fig:overview}
\end{figure*}

\subsection{Overview}
Given a sequence of limited FOV ground images, our goal is to geo-localize an aerial image from a reference database from where this sequence was taken. To achieve this goal, we considered geo-localization as a retrieval task similar to many other previous works~\cite{SAFA,Vigor,cvmnet,Vo,lin2015learning,liu2019lending,comingDown}. Specifically, we denote extracted features from a geo-tagged aerial image as $F_{sat}$ and extracted aggregated features from a sequence of ground-level images as $F_{grd}$. By evaluating the distance between $F_{sat}$ and $F_{grd}$, we can find the most similar aerial image from a database of aerial images. To extract features from a sequence of ground-level images, we introduce an end-end model to extract the sequential spatio-temporal features. We use VGG16~\cite{vgg} to extract the spatial features from each image and then pass those features to a novel Temporal Feature Aggregation Module (TFAM) to capture the temporal information. The spatio-temporal features are then aggregated into a single feature for retrieval. Furthermore, to generalize the proposed method for different sequence lengths, a sequential dropout (SD) scheme is implemented.  Fig.~\ref{fig:overview} provides an overview of the proposed approach. In the next sections, we describe TFAM in more detail followed by an introduction to the sequential dropout (SD) scheme in Section~\ref{sec::seq_drop}. Finally, in Section~\ref{sec::training}, we describe the training objectives.


\subsection{Temporal Feature Aggregation Module}
\label{sec::TFAM}
To explore the benefits of the sequential images, we introduce TFAM in cross-view sequence geo-localization. TFAM is inspired by the success of the transformers \cite{attention} in many computer vision problems~\cite{ViT,segmentationTrans,Regmi_2021_ICCV}. The multi-head self-attention mechanism is the key component that enabled transformers to capture correlations between sequential data elements at any distance. Similar to the transformer~\cite{attention}, TFAM also employs a multi-head self-attention mechanism to capture contextual information from a sequence of images.

Consider a sequence of images as $\mathcal{P} \in \mathbb{R}^{T\times W \times H \times C}$ where $T, W, H, C$ are the number of images in a sequence, image width, image height, and image channel respectively. We choose the VGG16 backbone to extract embedding features for each image in the sequence to have a fair comparison with the baseline methods~\cite{SAFA,Vigor}. 
A feature vector $F' \in \mathbb{R}^{T \times D}$ is obtained by concatenating each image's feature along the temporal axis, where $D$ is the dimension of the output of the backbone feature extractor. Similar to the original transformer~\cite{attention}, before the multi-head self-attention layer, a sinusoidal positional encoding $E_{pos} \in \mathbb{R}^{T \times D}$ is added to the extracted feature embeddings $F'$ to preserve the order of the temporal information which is shown in Equation~\ref{equation::position}.
\begin{equation}
    \Tilde{F} = F' + E_{pos}.
\label{equation::position}
\end{equation}
By feeding the feature embeddings to the Multi-head self-attention layer, each embedding vector is projected into three sub-spaces as $Q_i=\Tilde{F}W^Q_i,$ $ K_i=\Tilde{F}W^K_i,$ $V_i=\Tilde{F}W^V_i$ representing the query, key, value respectively and $i$ is the index of the head which we will describe later. Note that  $W^Q_i \in \mathbb{R}^{D \times \frac{D}{N_{head}}}, W^K_i \in \mathbb{R}^{D \times \frac{D}{N_{head}}}, $ and  $W^V_i \in \mathbb{R}^{D \times \frac{D}{N_{head}}}$ are three projection matrices and $N_{head}$ is the number of heads. The attention mechanism can be written as follows:
\begin{equation}
    head_i = softmax(\frac{Q_iK_i^T}{\sqrt{D}})V_i.
\end{equation}
To fully explore the contextual information in the temporal domain, we use an approach similar to~\cite{attention} and concatenate the value of the multiple heads and projected them to the output space using a projection matrix $W^O \in \mathbb{R}^{N_{head} D \times D}$,
\begin{equation}
    F_{aggregated} = Concat(head_1, head_2, ..., head_{N_{head}}) W^O.
\end{equation}
By stacking $N$ TFAM modules, our model can extract more refined feature representations. Finally, the features from the last TFAM, $F_{aggregated} \in \mathbb{R}^{T \times D}$ has the same input shape as the embedding vector $F$. The resulting features are then averaged using an average pooling layer on the temporal axis to obtain a one-dimension vector for the retrieval task as follows:
\begin{equation}
  F_{grds}=average\_pool(F_{aggregated}).  
\end{equation}

\subsection{Adaptive Sequence Length}
\label{sec::seq_drop}
TFAM introduced in the previous section works well on sequences that have a fixed length $T$. However, during inference in real-world settings, it is not always possible to capture exactly $T$ frames in a sequence due to different hardware or software configurations (e.g. different sampling and capturing rate, signal loss, etc). To adapt TFAM to variant sequence lengths, we propose a sequential dropout (SD) scheme by modifying the TFAM algorithm. During training, a random binary mask $A \in \mathbb{R}^{T}$ is generated and fed to each TFAM in the model. For each index $x$ at $A_x$, if $A_x=0$, it means that the feature at index $x$ in $\Tilde{F}_x$ is omitted. Otherwise, the TFAM operates normally on this feature. By setting $K_{i,x}$ to a zero vector at index $x$, the attention value at head $i$ of index $x$ represented as $Q_{i}K_{i,x}^T$ also becomes a zero vector. In other words, all the query values would never interact with the key values of this feature at index $x$. Consequently, the embedding vector $\Tilde{F}_x$ at index $x$ of $\Tilde{F}$ is ignored by all other vectors during forward propagation and back-propagation. To generate the random mask $A$, we set a maximum number of dropout features $J$ in which $J < T$. In each training mini-batch, we uniformly sampled an integer $e$ between $[0,J]$ to represent the number of dropped features in this batch. To control the dropout rate during the training, we initialized all $A$ values with $1$, and randomly set $e$ elements in $A$ to $0$. The average pooling layer (mentioned in section~\ref{sec::TFAM}) only operates on the index of the temporal dimension of aggregated features $F_{aggregated}$ where the mask value is $1$. Note that to fully exploit the temporal information, our approach does need a fixed-length sequence during training, however, it employs SD to tackle variable-length sequence during testing. In our experiments, we found that this strategy not only helped the TFAM to produce a coherent representation but also regularized the model and achieved much higher performance.

\subsection{Training Objective}
\label{sec::training}

After extracting the aerial features $F_{sat}$ and ground-level features $F'$, $F'$ is
further refined using the proposed TFAM as described in~\ref{sec::TFAM}, and the aggregated ground-level features $F_{grd}$ are obtained. Finally, we deploy a metric learning objective to train the model using weighted soft margin triplet loss~\cite{cvmnet},
\begin{equation}
    \mathcal{L} = log(1+e^{\gamma(d_{pos} - d_{neg}})),
    \label{equation::loss}
\end{equation}
where $\gamma$ is a hyperparameter that controls the scale of the loss value. $d_{neg}$ and $d_{pos}$ are $L_2$ distances of unmatched and matched aerial-ground pairs. We employ $L_2$ normalization on $F_{sat}$ and $F_{grd}$ before calculating the distance. The goal of this loss function is to push the matched pairs closer while pushing unmatched pairs further.
\section{Experiments}
\label{sec:exp}

\noindent\textbf{Implementation Details \& Dataset:} The proposed method was implemented in PyTorch~\cite{pytorch} \footnote{Codes available at \url{https://gitlab.com/vail-uvm/seqgeo}}. We used VGG16~\cite{vgg} pretrained on ImageNet~\cite{imagenet} as backbones of our features extractors. The last two fully connected layers were removed for extracting features. 
We stacked $6$ TFAMs, each with $8$ heads, in our model. We adopted our proposed SD scheme with a maximum number of dropout features $J=6$ during training. During the testing, the frames can be dropped by setting the values at corresponding locations in $A$ to $0$.
Since our proposed method exploit sequence images for training, we cannot evaluate our method on existing cross-view geo-localization datasets. Instead, we chose to benchmark our proposed method on our dataset described in Section~\ref{sec:dataset}. The dataset is split into training and testing sets with $31,091$ and $7,772$ aerial-ground sequence pairs respectively. The training and testing datasets are geographically separated that no overlapping areas between these two datasets. These settings are applied to all the experiments in this section unless specified otherwise.

\noindent\textbf{Baseline Methods:} We compare our method with SAFA~\cite{SAFA} and VIGOR~\cite{Vigor} on our dataset. We chose SAFA~\cite{SAFA} as it achieved very competitive results on both CVUSA~\cite{CVUSA} and CVACT~\cite{liu2019lending} datasets. VIGOR~\cite{Vigor} also achieved outstanding performance on their proposed dataset in a one-to-many retrieval approach.
To adopt SAFA~\cite{SAFA} on our dataset, we trained SAFA~\cite{SAFA} on center ground-level images with their corresponding aerial images on our proposed dataset with configurations reported in its original paper. Thus, to make the comparison fair, we initialized SAFA with pre-trained weights on the CVUSA dataset. To be noticed, we did not apply the polar transformation in SAFA~\cite{SAFA} to keep a fair comparison with other methods. To train VIGOR, we set the center ground-level image as a `positive' sample and the others are `semi-positive' samples as defined in their original paper. To enable SAFA and VIGOR to work on sequences of images, we feed each ground-level image in the sequence separately and average the final feature vectors for all the images.\\
\noindent\textbf{Evaluation Metrics:} Similar to the previous works~\cite{SAFA,cvmnet,liu2019lending,Vigor}, we use the recall accuracy at top-K (R@K) for evaluating the performance. Given a query sequence, if the ground truth aerial image ranks in the first $K$ most similar aerial images, it is considered to be a `correct' query. 

\subsection{Quantitative comparison}

\begin{table}[!t]
\centering
\begin{tabular}{c|cccc}
\toprule
& R@1    & R@5    & R@10   & R@1\% \\ \midrule
VIGOR~\cite{Vigor}                               & 0.54\% & 2.52\% & 4.48\% & 18.55\%\\
$\text{SAFA}^{\dag}$~\cite{SAFA}                        & 0.68\% & 2.92\% & 5.06\% & 21.81\% \\
SAFA~\cite{SAFA}                      & 0.63\% & 2.83\% & 5.03\% & 21.51\% \\ \hline
Ours w/o SD                               & 1.39\%       &\textbf{6.50}\%       &\textbf{10.45}\%         & 32.42\% \\
Ours w/ SD                               & \textbf{1.80\%}       &6.45\%     &10.36\%         & \textbf{34.38\%} \\  \bottomrule
\end{tabular}%
\caption{Comparison between our methods with SD and without SD, SAFA and VIGOR methods. $\dag$ indicates testing on single center ground image as query.}
\label{table::main_result}
\end{table}

\begin{figure}[!t]
    \centering
    \includegraphics[width=0.45\textwidth]{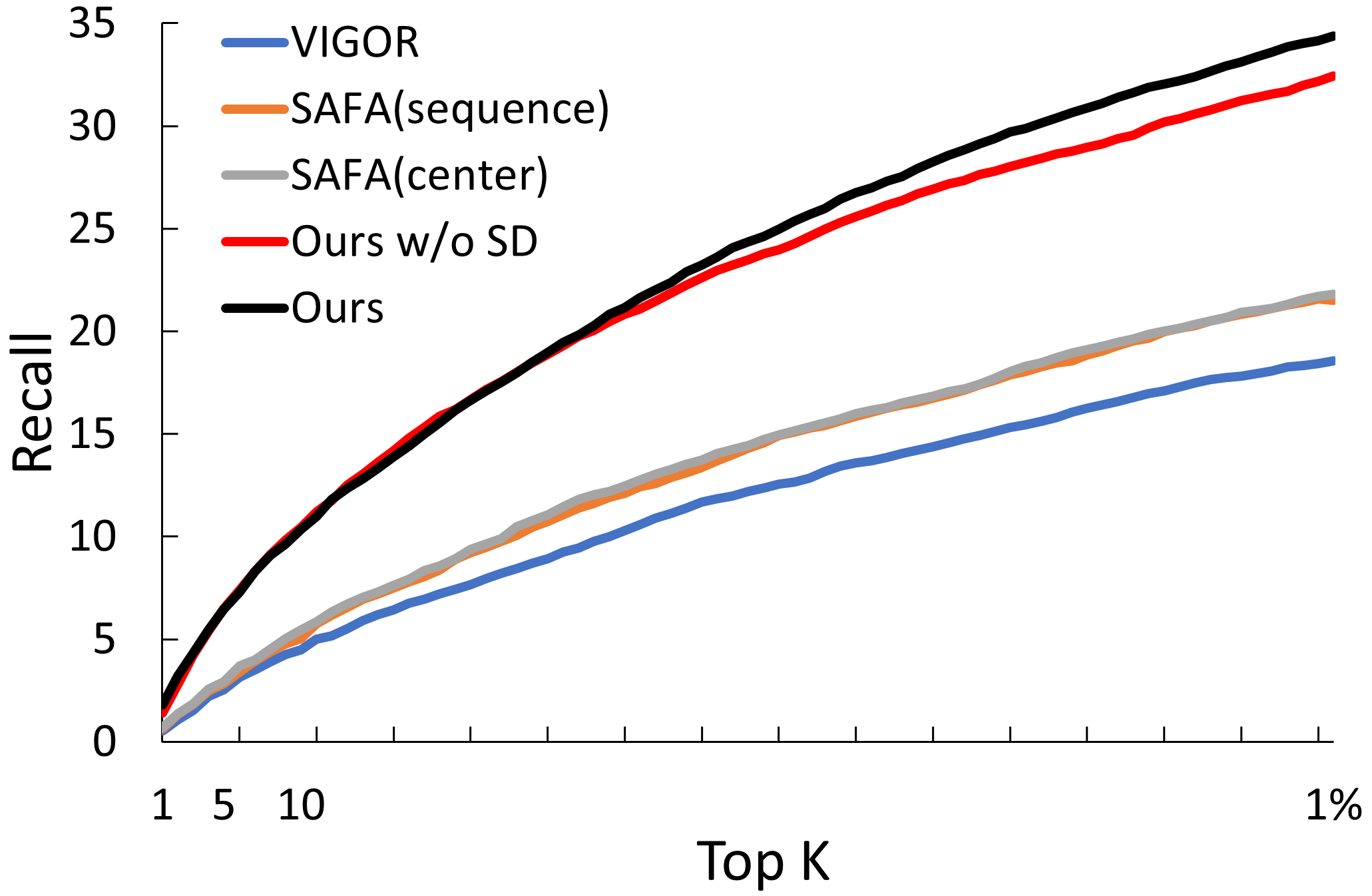}
    \caption{Recall rates of our methods vs baseline methods. The results demonstrate that both methods trained with SD and without SD outperform the baseline methods.}
    \label{fig:recall_vs_topk_plot}
\end{figure}

\begin{figure}[!t]
    \centering
    \includegraphics[width=0.45\textwidth]{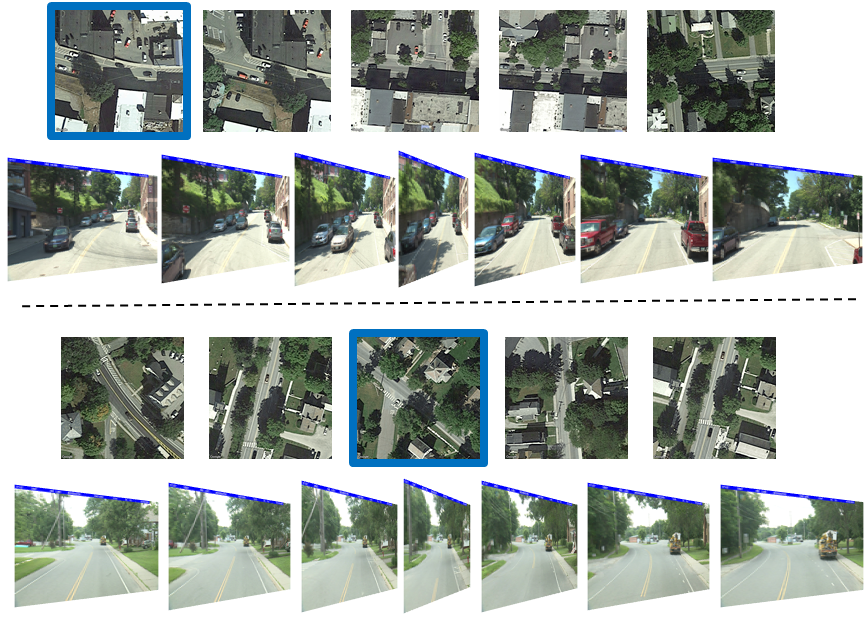}
    \caption{Two randomly selected retrieval results. The top row shows top-5 retrieved aerial images in descending order. The bottom row is the query sequence. The aerial images with blue border are the ground truth.}
    \label{fig: sample_retrive}
\end{figure}

Our main results are reported in Table~\ref{table::main_result}. SAFA(center) indicates that the SAFA model was tested on the center ground-level image only. SAFA(sequence) means that the SAFA model was tested on the whole sequence of ground-level images after averaging the features. We also provide the results from our method without SD. It can be seen that our method outperforms the baseline methods by a large margin. We also observe that our method performs much better with SD in both top-1 and top-1\% recall. The model without SD is slightly better than the model with SD on top-5 and top-10 recalls, but it is a slight margin as shown in the recall vs top-K graph in Fig.~\ref{fig:recall_vs_topk_plot}. Two randomly selected aerial-ground sequence pairs predicted by our model training with SD from our test set are visualized in Fig.~\ref{fig: sample_retrive}. In the top two rows of Fig.~\ref{fig: sample_retrive}, the ground truth image was successfully predicted as the most similar one. It is worth noticing that the second, third and fourth aerial images share most of their appearance with the top-1 image. In the bottom two rows of Fig.~\ref{fig: sample_retrive}, although our model fails to predict the ground truth in the first place, we can see that visually the top-1 prediction is very similar to the ground truth.

\subsection{Ablation Studies}

\begin{table}[!t]
    \centering
    \begin{tabular}{cc|cccc}
    \toprule
    T & H & R@1    & R@5    & R@10    & R@1\%   \\ \midrule
    0 & 0 & 0.91\% & 4.49\% & 7.98\% & 26.69\% \\
    2 & 2 & 1.45\% & 6.22\% & 10.02\% & 31.84\% \\
    4 & 2 & 1.40\% & 6.34\% & 10.31\% & 32.97\% \\
    4 & 4 & 1.51\% & 6.27\% & 10.51\% & 32.93\% \\
    6 & 4 & 1.59\% & 6.02\% & 9.88\%  & 32.14\% \\
    6 & 8 & \textbf{1.80}\% & \textbf{6.45}\% & \textbf{10.36}\% & \textbf{34.38}\% \\ \bottomrule
    \end{tabular}%
    \caption{Ablation study on the number of heads and TFAMs. `T' is short for number of TFAMs and `H' is short for number of  Heads. $J$ is fixed to 6.}
    \label{table::ablation_heads_tfams}
\end{table}

\begin{table}[!t]
    \centering
    \begin{tabular}{c|cccc}
    \toprule
           & R@1   & R@5   & R@10   & R@1\%      \\ \midrule
    $J=1$ &1.40\%      &6.08\%      &9.45\%       &31.89\% \\
    $J=3$ &1.51\%      &6.64\%      &10.57\%       &34.34\% \\
    $J=5$ &1.63\%      &6.41\%      &10.49\%       &34.40\% \\
    $J=6$ &\textbf{1.80}\%      &\textbf{6.45}\%      &\textbf{10.36}\%       &\textbf{34.38}\% \\ \bottomrule
    \end{tabular}%
    \caption{Ablation study on the maximum number of masked frames $J$. The model is fixed to 6 TFAMs with 8 heads.}
    \label{table:ablation_mask_recall}
\end{table}

\begin{table}[!t]
    \centering
    \begin{tabular}{c|cccc}
    \toprule
         BackBone & R@1   & R@5   & R@10   & R@1\%  \\ \midrule
         VGG16~\cite{vgg}&1.80\% & 6.54\% & 10.36\% & 34.38\% \\
         ResNet18~\cite{resnet} & 1.58\% & 5.98\% & 10.14\% & 33.83\% \\
         ResNet34~\cite{resnet} & 1.71\% & 7.01\% & 11.67\% & 38.16\% \\
         ResNet50~\cite{resnet} & 2.07\% & 8.12\% & 13.16\% & 40.10\% \\
         \bottomrule
         
    \end{tabular}
    \caption{Comparison between different backbones of the proposed model.}
    \label{tab:backbone}
\end{table}

To evaluate the effectiveness of our proposed models, we conducted ablation experiments. We study the effectiveness of the TFAM modules, SD, and number of heads in the TFAM module as reported in Table~\ref{table::ablation_heads_tfams} and Table~\ref{table:ablation_mask_recall}. We observe that the model which has $6$ TFAMs with $8$ heads for each multi-head self-attention layer and random dropout of a maximum of $6$ images achieved the best results among all the configurations. Moreover, we evaluate performance of our model under different backbones in Table~\ref{tab:backbone}. To be noticed, our model with ResNet50~\cite{resnet} can achieve $40\%$ on R@1\%. But to keep a fair comparison with baseline methods, we still use VGG16~\cite{vgg} as the backbone.


\begin{figure}[!t]
\centering
\begin{subfigure}[b]{0.24\textwidth}
\centering
\includegraphics[width=\textwidth]{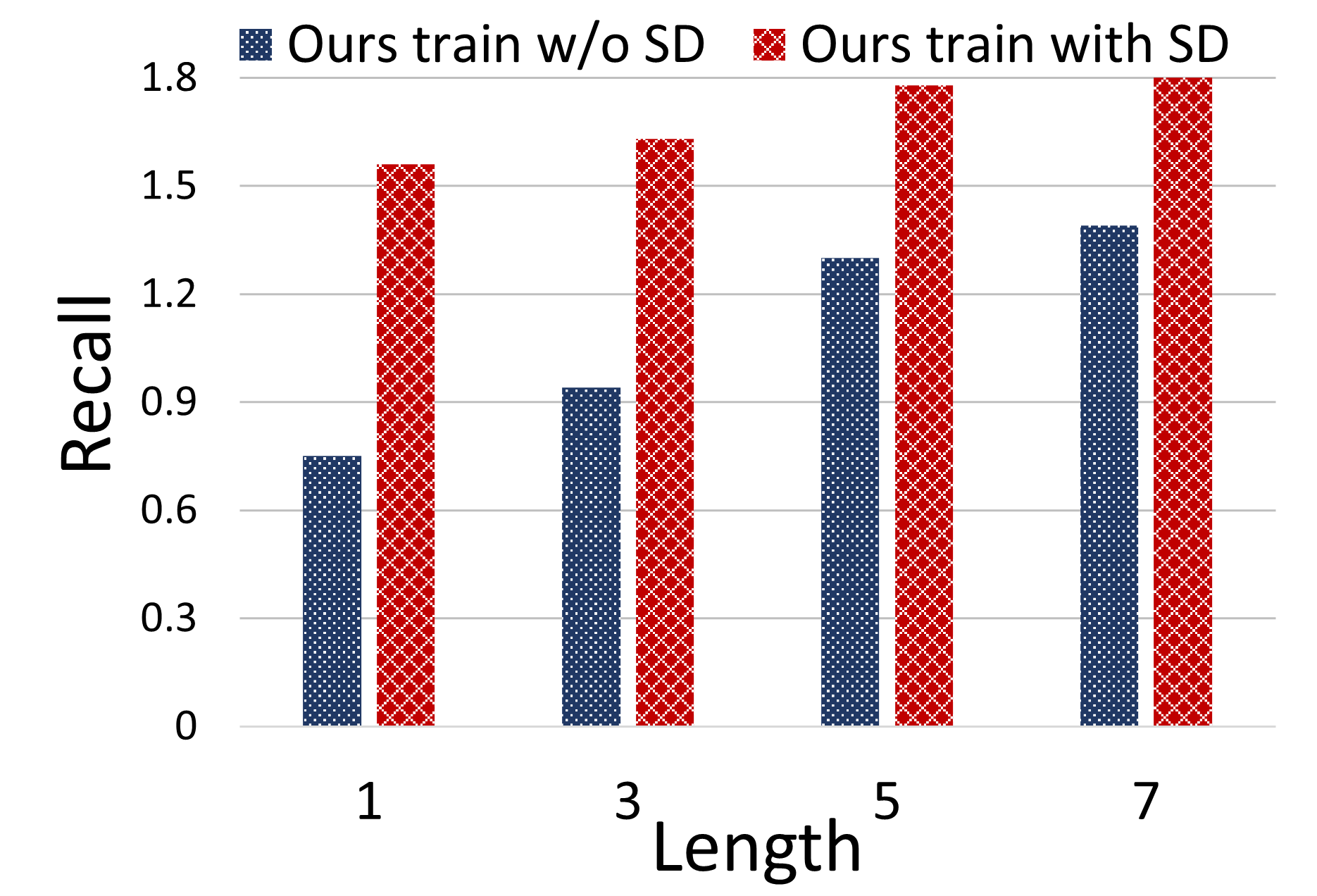}
\caption{Recall@1}
\end{subfigure}
\hspace{-8pt}
\begin{subfigure}[b]{0.24\textwidth}
\centering
\includegraphics[width=\textwidth]{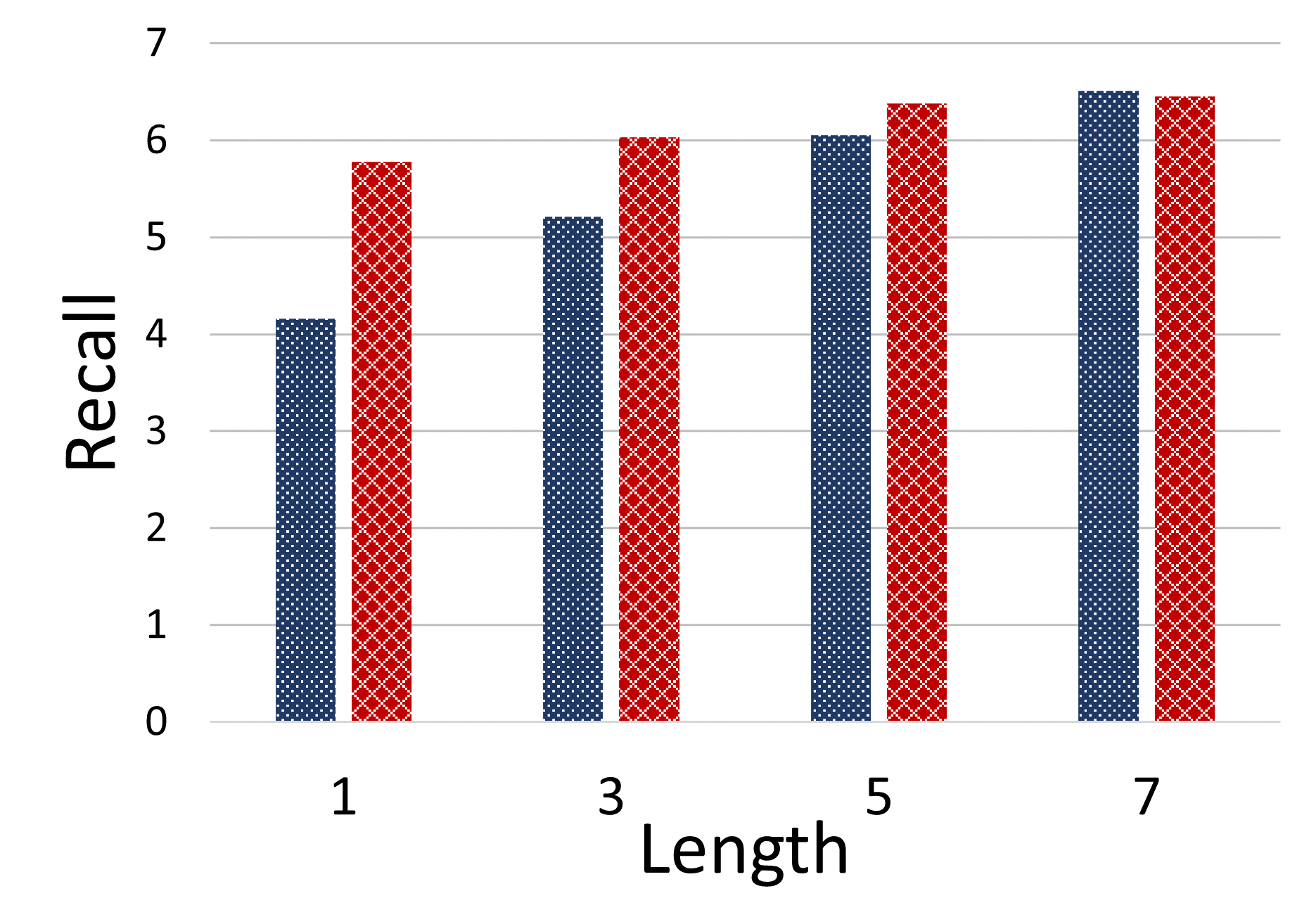}
\caption{Recall@5}
\end{subfigure}
\hspace{-8pt}
\begin{subfigure}[b]{0.24\textwidth}
\centering
\includegraphics[width=\textwidth]{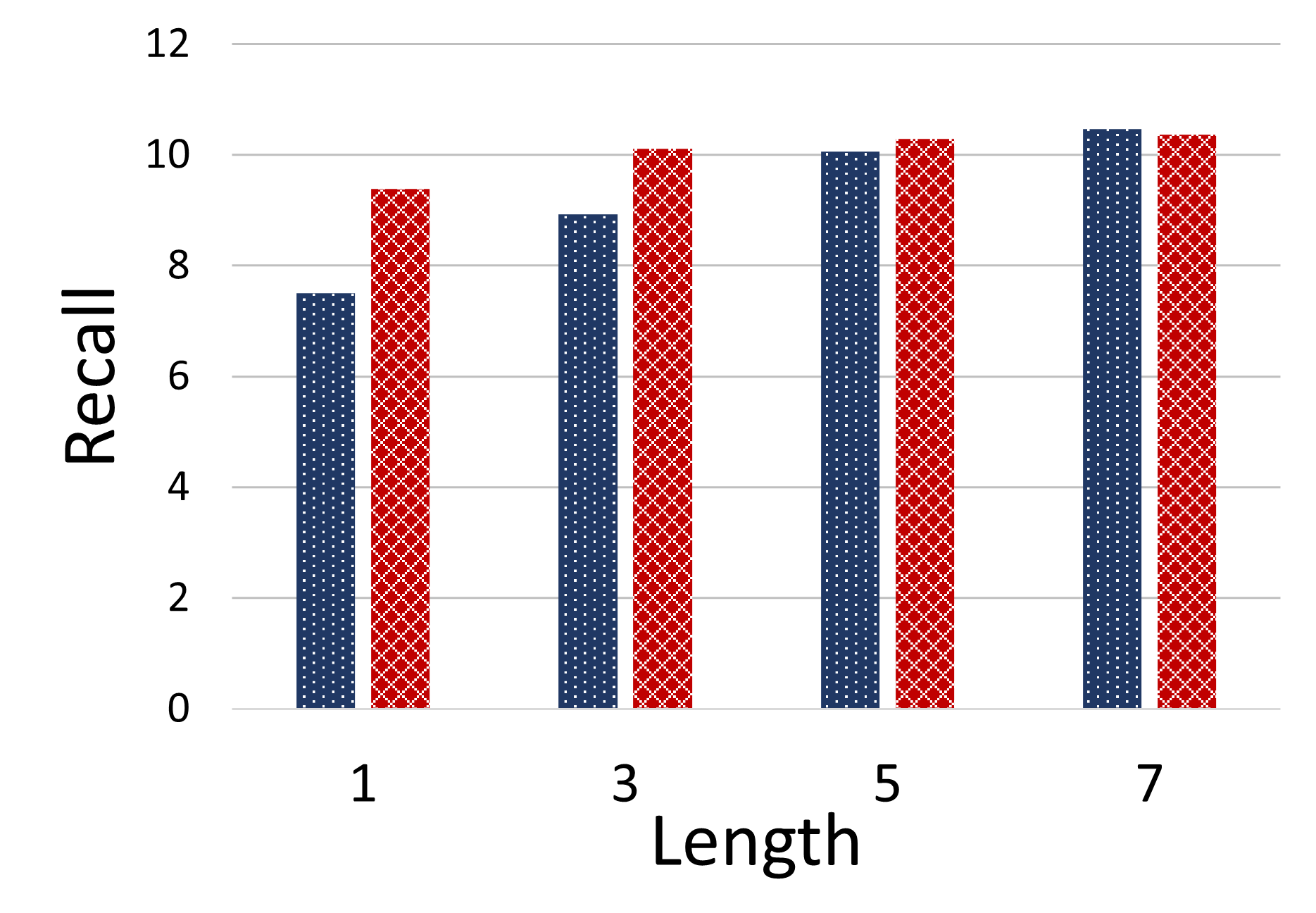}
\caption{Recall@10}
\end{subfigure}
\hspace{-8pt}
\begin{subfigure}[b]{0.24\textwidth}
\centering
\includegraphics[width=\textwidth]{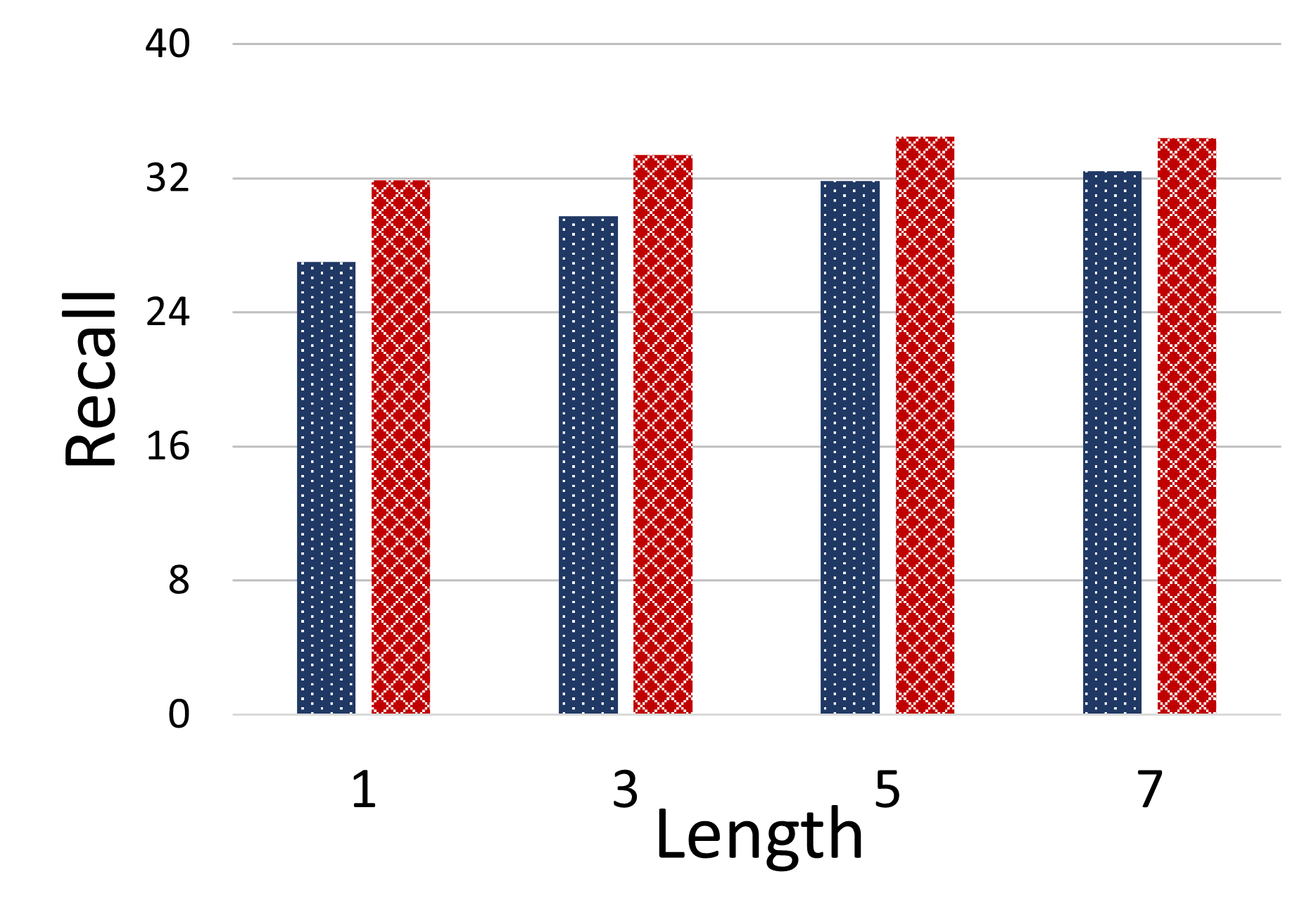}
\caption{Recall@1\%}
\end{subfigure}
\caption{Comparison between variant sequence lengths simulated by SD in the testing phase for two models trained with (red) SD and without (black) SD. The results demonstrate that although trained using fixed-length sequences, the proposed SD enables our method to predict coherent feature representation compared with training without SD.}
\label{fig:vary_length}
\end{figure}

\subsection{Variant Sequence Lengths}
In real-world scenarios, the ground-level sequences could have different numbers of images. Given that our model has been trained with the SD scheme, in this experiment, we vary the number of ground-level images in a sequence at the inference time by modifying the value of the SD mask $A$. We compare our model with and without SD. To simulate the worst possible real-world scenarios, we started by dropping the first $6$ images and only leaving the last $1$ image in the sequence as the last image has the smallest visible overlap area with the aerial image. We then test by dropping the first $4$ images and $2$ images respectively. The results in Fig.~\ref{fig:vary_length} demonstrated that our model with SD outperformed the model trained without it in most of the cases which proves that SD improves the model performance and feature coherency on variable length sequences. Noticeably, even ground sequence has only one single image during testing, our model trained with SD significantly outperforms SAFA~\cite{SAFA} which was trained with the center image as query (Table~\ref{table::main_result}) as observed from Fig~\ref{fig:vary_length}.
\section{Conclusion, Limitation, and Future Work}
\label{sec:conclusion}
In this paper, we put forward the first cross-view geo-localization method that operates on sequences of limited FOV images. To aggregate the temporal features, we proposed a TFAM module that leveraged the multi-head self-attention mechanism to fuse information from a sequence of images. Although we used fixed-length sequences during the training phase, we simulated variant length sequences using our proposed sequential dropout method that regularizes our model to have a coherent feature representation. This also helps our model to tackle ground-level sequences with different lengths during the testing phase. We contributed to the vision community a novel large-scale cross-view \textit{sequence} geo-localization dataset. Our extensive experiments demonstrated the effectiveness of different components of the proposed approach, robustness on variable-length input sequences, and state-of-the-art results against several competitive cross-view geo-localization methods.

One limitation of our proposed method is that the maximum length of the ground-level image sequence is constrained by the size of aerial images. Exploring methods that can geo-localize long sequences spanning multiple aerial images is one future research direction.


{\small
\bibliographystyle{ieee_fullname}
\bibliography{references}
}
\clearpage
\newpage

\appendix

\section*{Supplementary Material} 

In this supplementary material, we are providing additional information for the following items:
\begin{itemize}
    \item The availability of panoramas and limited Field-Of-View (FOV) images.
    \item Dataset coverage map
    \item More implementation details
    \item More details about the baseline methods discussed in the main paper.
    \item Comparison of the number of trainable parameters between the proposed model and baseline methods
    \item The availability of our proposed dataset and code for the public.
    \item More samples from our proposed dataset.
    \item More qualitative results predicted by our proposed model.
   \end{itemize}


\begin{figure*}[!ht]
     \centering
     \begin{subfigure}[b]{0.49\textwidth}
         \centering
         \includegraphics[width=\textwidth]{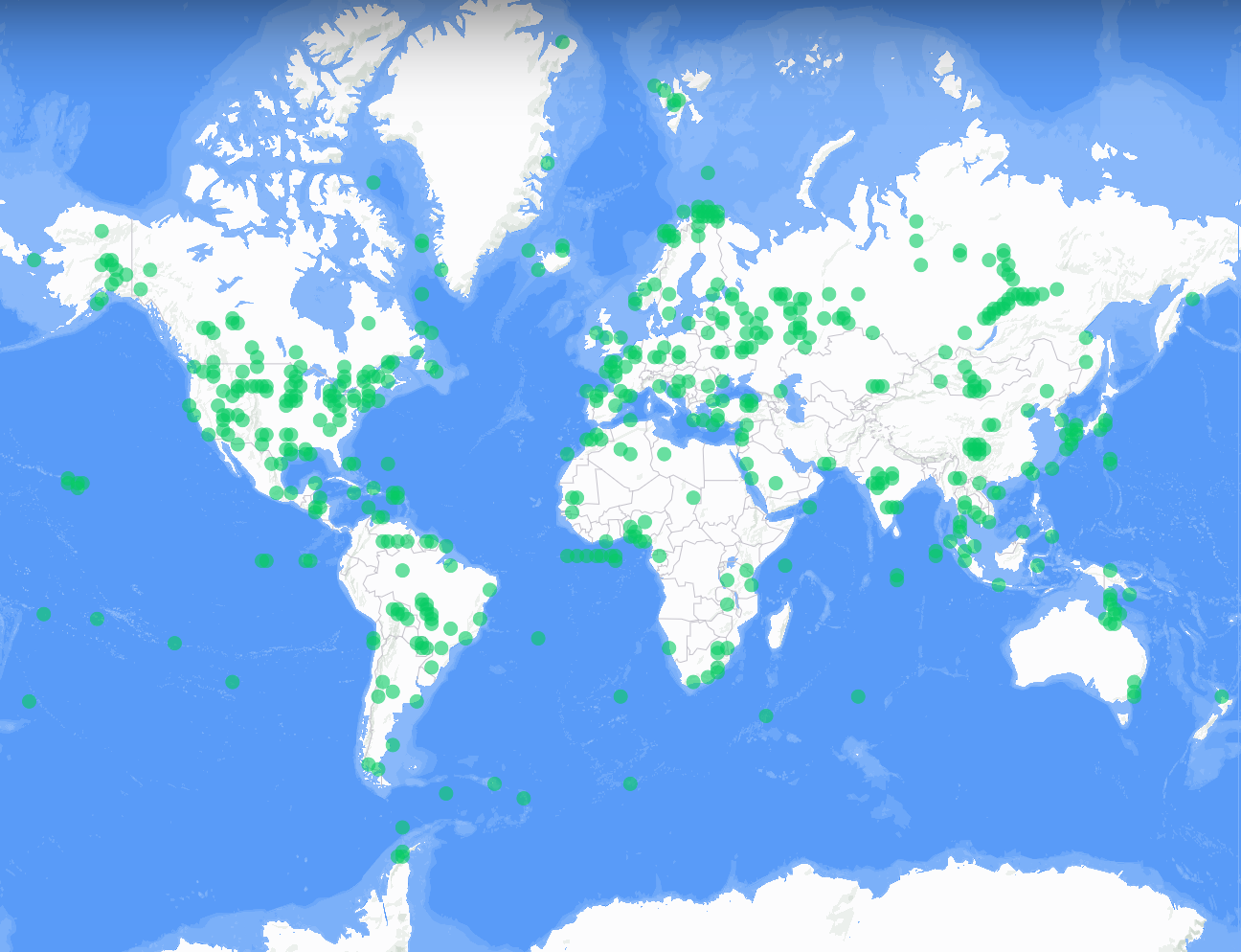}
         \caption{}
         \label{fig:pano_cover_2}
     \end{subfigure}
     \begin{subfigure}[b]{0.49\textwidth}
         \centering
         \includegraphics[width=\textwidth]{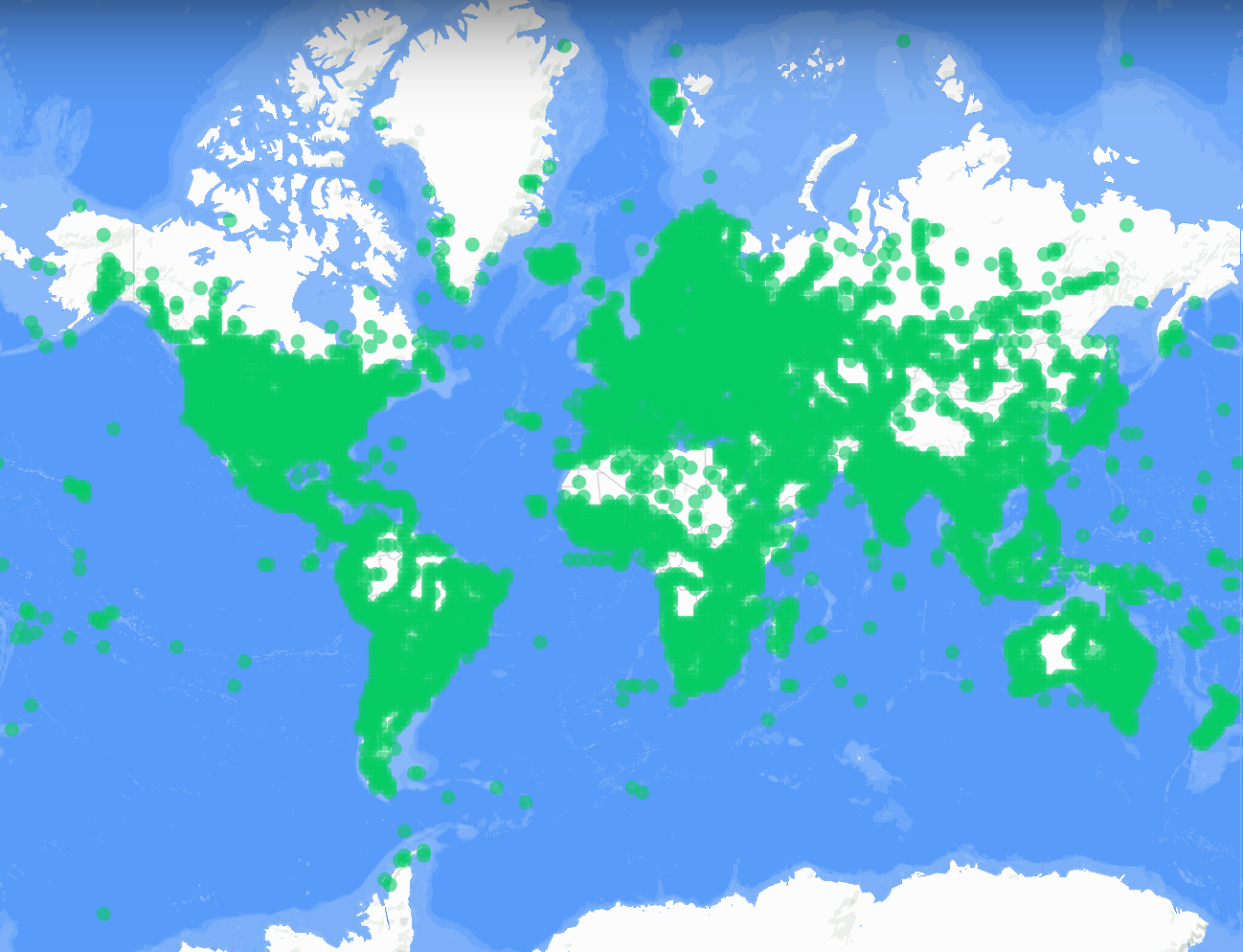}
         \caption{}
     \label{fig:lfov_cover}
     \end{subfigure}
     \caption{Comparison of coverage area (green lines) of user uploaded street view images between panoramic (a) and limited FOV images (b) on Mapillary~\cite{mapillary}.}
     \label{fig:comparison}
\end{figure*}

\section{Panorama vs Limited FOV images}
As we discussed in our main paper, limited FOV images are more popular and common than panoramas. To highlight the difference, we presented the coverage areas of limited FOV images and panoramas from Mapillary~\cite{mapillary} in Fig~\ref{fig:comparison}. Mapillary~\cite{mapillary} is one of the largest crowdsourcing platforms for sharing geotagged photos. As of 2018, Mapillary~\cite{mapillary} hosted 422 million images across the world. As observed from Fig~\ref{fig:comparison}, the coverage area of limited FOV images (Fig.~\ref{fig:lfov_cover}) on Mapillary is substantially greater than the coverage area of panoramas (Fig.~\ref{fig:pano_cover_2}), especially in some developing areas such as Middle East , Africa and south America. We refer this to the complexity of capturing panoramic images which they need special and expensive cameras. To this end, using sequences of limited FOV images as the query is much more practical than using panoramas as the query in cross-view geo-localization.

\section{More implementation details}

Our model was trained in an end-to-end manner using Adam~\cite{adam} with weight decay of $10^{-6}$ for $50$ epochs on a single Nvidia V100 GPU. The learning rate is set initially to $10^{-5}$ and decayed linearly to $5 \times 10^{-7}$ after $30$ epochs. We set the $\gamma$ in Equation 5 of main paper to $10$. We set the ground sequence length $T=7$ which is suitable for our dataset. We used the exhaustive mini-batch strategy~\cite{Vo} to construct the triplet pair with batch size set to $24$.

\section{Baseline Methods}
\label{sec:baseline}

We employed two baseline methods for comparison, SAFA~\cite{SAFA} and VIGOR~\cite{Vigor}. For SAFA~\cite{SAFA}, we adopted their original code. \footnote{\url{https://github.com/shiyujiao/cross\_view\_localization\_SAFA}} SAFA trained only on the center images of each sequence. For fair comparison, SAFA has been initialized with weights pretrained on CVUSA~\cite{CVUSA} dataset then trained on our dataset. We used same hyperparameters reported in SAFA's original paper~\cite{SAFA} and fine-tuned the model for $10$ epochs. For VIGOR~\cite{Vigor}, we used their  code\footnote{\url{https://github.com/Jeff-Zilence/VIGOR}} for training. Similar to SAFA, we trained their model from all images in the sequences by setting the center ground-level image to a `positive' sample and the others are `semi-positive' samples as defined in their original paper. We set the hyperparameters as reported in original VIGOR paper~\cite{Vigor} and followed their exact procedures for training.

\section{Dataset Availability and Anonymity}
\label{section::dataset}

Our proposed dataset is composed of two parts, ground-level image sequences and satellite imagery as explained in the main paper. Our ground-level images are public images collected by Vermont Agency of Transportation\footnote{\url{https://vtrans.vermont.gov/}}. The private information of all ground-level images has been anonymized. These images will be shared publicly. Our satellite images came from Google Maps. Following Google Maps Platform Terms of Service\footnote{https://cloud.google.com/maps-platform/terms}, we will make our dataset available for research purposes only. We will follow existing datasets, such as VIGOR~\cite{Vigor}, to distribute the collected dataset upon request.

\section{Dataset Coverage Map}

\begin{figure}[!ht]
    \centering
    \includegraphics[width=0.45\textwidth]{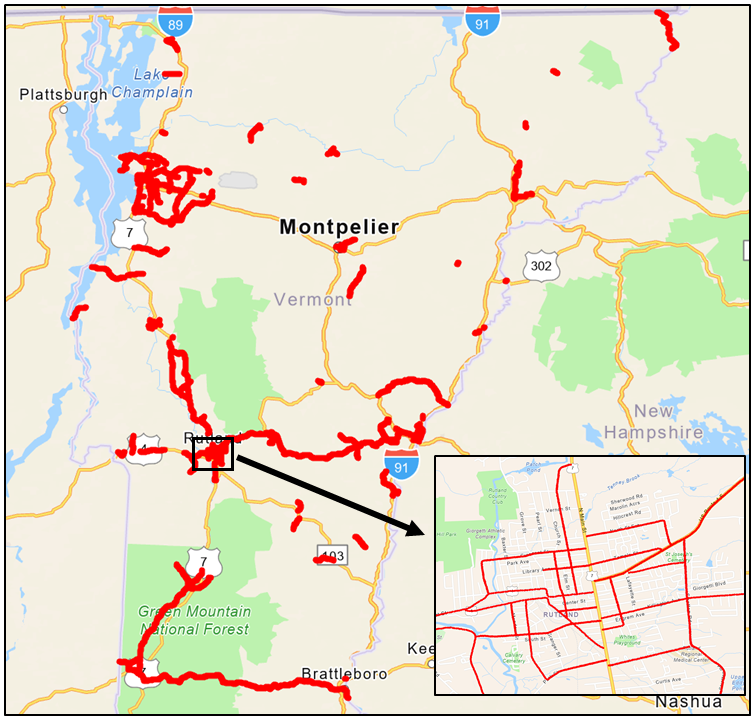}
    \caption{The coverage map of the proposed dataset. The coverage area is indicated by red lines.}
    \label{fig:coverage}
\end{figure}

To better visualize the diversity of the proposed dataset, we visualize the coverage area in Fig.~\ref{fig:coverage}. As indicated by the coverage map, our dataset includes both suburban and urban areas in Vermont, US which cover most scenarios on the roads.

\section{Comparison of parameters}

\begin{table}[!t]
    \centering
    \footnotesize
    \begin{tabular}{c|c|cccc} \toprule
        Method & Parameters & R@1 & R@10 & R@1\% \\ \midrule
        VIGOR~\cite{Vigor} &
        395M & 0.54\% & 4.48\% & 18.55\%\\
        $\text{SAFA}^{\dag}$~\cite{SAFA} & 319M & 0.68\% & 5.06\% & 21.81\% \\
        Ours w/ VGG16~\cite{vgg} & 2.9G & 1.80\% & 10.36\% & 34.38\% \\
        Ours w/ Res50~\cite{resnet} & 775M & 2.07\% & 13.16\% & 40.10\% \\
        Ours w/ Res34~\cite{resnet} & 240M & 1.71\% & 11.67\% & 38.16\% \\
        Ours w/ Res18~\cite{resnet} & 161M & 1.58\%  & 10.14\% & 33.83\% \\ \bottomrule
    \end{tabular}
    \caption{Comparison between our proposed methods with different backbones and baseline methods. $\dag$ indicates testing on single center ground image as query.}
    \label{tab:parametes}
\end{table}

In this section, we present the comparison of trainable parameters between the proposed model with different backbones and baseline methods in Table~\ref{tab:parametes}. Our model with VGG16~\cite{vgg} is larger than the baselines. This is because the output dimension of VGG16 is $4096$. As a result, we need wider TFAMs to handle this large latent vector. When we switch to ResNet~\cite{resnet} as backbone, the number of parameters is significantly less than VGG~\cite{vgg} as backbone. This is because the dimension of output of ResNet50 is $2048$. For ResNet34 and ResNet18, the dimension of output is only $512$ which cause these two models are even smaller than baselines. However, despite of the backbones, the proposed model is constantly outperforms the baseline methods. For a fair comparison with baseline methods, we choose VGG16 as the backbone in the main script.

\section{More Dataset examples}
\begin{figure*}[!ht]
    \centering
    \includegraphics[page=3,width=0.99\textwidth]{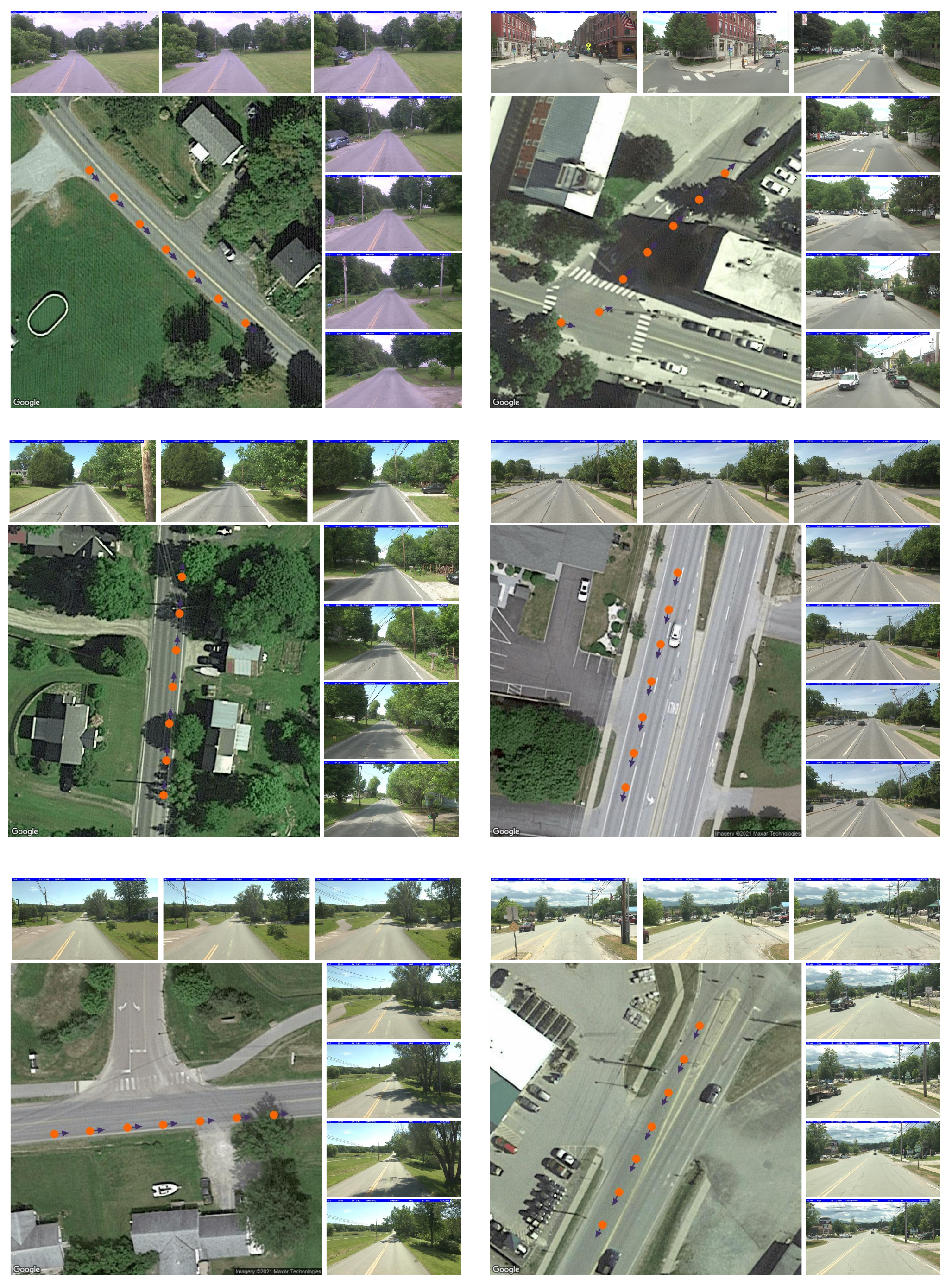}
    \caption{Six randomly sampled satellite and ground sequence pairs from our dataset.}
    \label{fig:dataset_more}
\end{figure*}

In this section, we provided 6 randomly sampled satellite and ground sequence pairs from our proposed dataset as shown in Fig.~\ref{fig:dataset_more}. As shown in Fig.~\ref{fig:dataset_more}, our dataset covers diverse locations, urban, suburban, and rural areas which we discuss in detail in our main script.

\section{More Qualitative Results}
\label{section::more_results}
\begin{figure*}[!ht]
    \centering
    \includegraphics[page=1,width=0.99\textwidth]{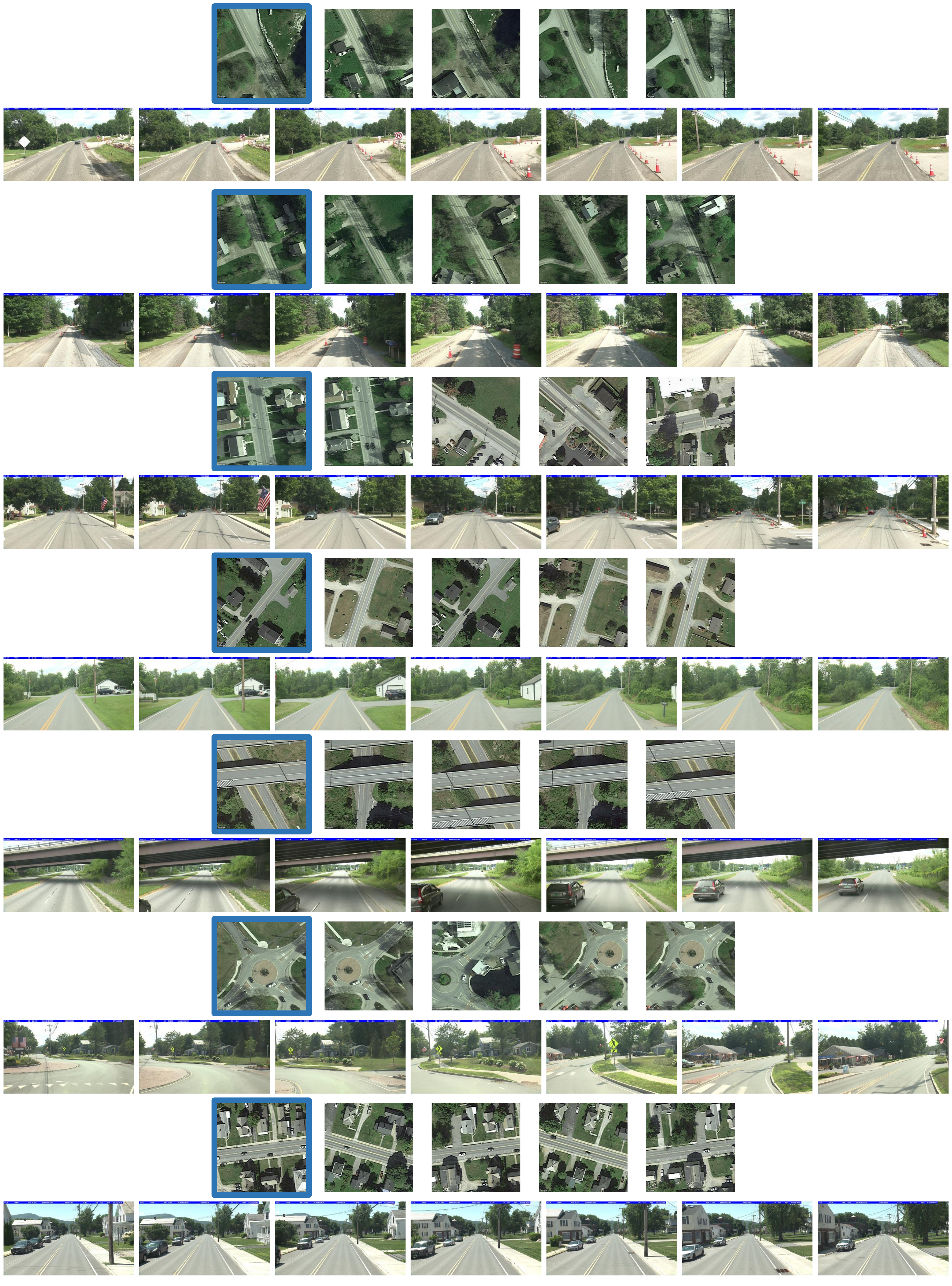}
    \caption{Samples been correctly predicted as top-1 by our model.}
    \label{fig:top1}
\end{figure*}

\begin{figure*}[!ht]
    \centering
    \includegraphics[page=2,width=0.99\textwidth]{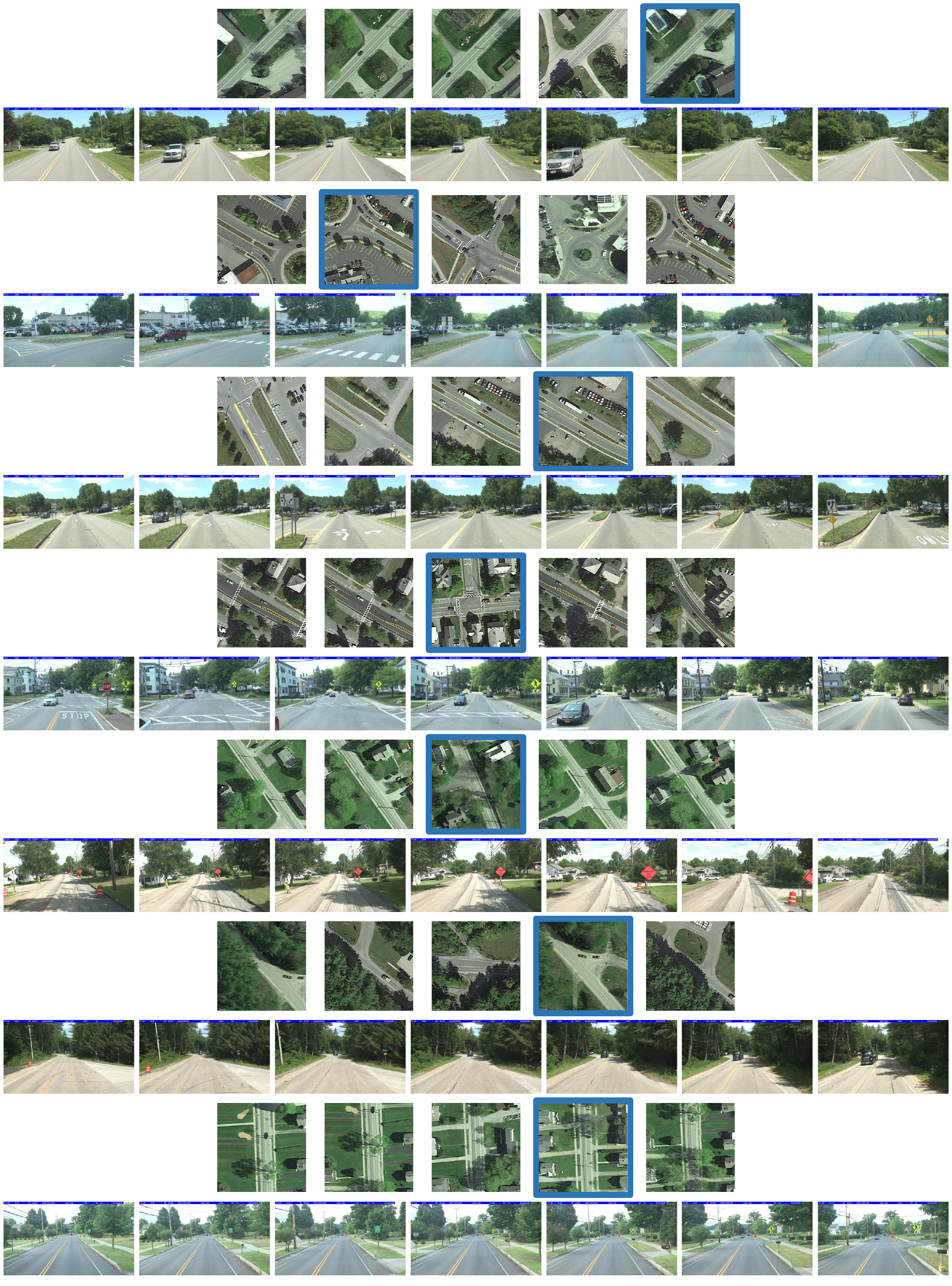}
    \caption{Samples been correctly predicted as top-5 by our model.}
    \label{fig:top5}
\end{figure*}

In this section, we provided more retrieval examples. Fig.~\ref{fig:top1} shows correct top-1 examples predicted by our model and Fig.~\ref{fig:top5} shows top-5 retrieval examples. Each figure shows pairs of satellite and ground images ordered from top to bottom. For each pair, the bottom row is the query ground-level sequence and the upper row is the predicted top-5 satellite images ranked in descending order from left to right. The satellite images with blue boarder are the ground truth.

\end{document}